%% file: bare_conf.tex
\documentclass[conference]{IEEEtran}
\usepackage{eurosym}

\usepackage{cite}

\ifCLASSINFOpdf
  \usepackage[pdftex]{graphicx}
  \usepackage{import}
\else
\fi
\usepackage[english]{babel}
\usepackage[utf8]{inputenc}
\usepackage[]{caption}
\usepackage{subcaption}
\usepackage{placeins}
\usepackage{color}
\usepackage{mathtools}
\usepackage{placeins}
\usepackage{relsize}
\usepackage{amssymb}
\usepackage{units}
\usepackage{amsmath}
\usepackage{float} 		

\usepackage{tikz}
\usepackage[color=black,opacity=1,angle=0,scale=1]{background}
\backgroundsetup{
contents={
\begin{tikzpicture}
\node at (current page.center) [align=center] {\textcopyright 2018 IEEE. Personal use of this material is permitted. 
\\ Permission from IEEE must be obtained for all other uses, in any current or future media, including reprinting/republishing this material for advertising or promotional purposes, 
\\ creating new collective works, for resale or redistribution to servers or lists, or reuse of any copyrighted component of this work in other works.
\\ DOI: 10.1109/IVS.2018.8500667};
\end{tikzpicture}},
placement=bottom,
scale=0.6,
vshift=20
}

\begin{document}
%
\title{\vspace{0.5cm} 
Optimization of Velocity Ramps with Survival Analysis for Intersection Merge-Ins}


\author{\IEEEauthorblockN{Tim Puphal$^{1\dagger}$, Malte Probst$^{1\dagger}$, Yiyang Li$^{2}$, Yosuke Sakamoto$^{2}$ and Julian Eggert$^{1\dagger}$}
\IEEEauthorblockA{$^{1}$ Honda Research Institute (HRI) Europe, 
	Carl-Legien-Str. 30, 63073 Offenbach, Germany \\
	Email: {\tt\small \{tim.puphal, malte.probst, julian.eggert\}@honda-ri.de} \\
	$^{2}$ Honda Innovation Lab (HIL) Tokyo, Honda R\&D Co., Ltd. 5-3-1 Akasaka, 107-6327 Tokyo, Japan \\
	Email: {\tt\small \{Yiyang\_Li, Yosuke\_Sakamoto\}@n.t.rd.honda.co.jp} \\
	$\dagger$ The authors contributed equally to this work}}


%


\maketitle

\begin{abstract}
We consider the problem of correct motion planning for 
T-intersection merge-ins of arbitrary geometry and vehicle density. A merge-in support system has to estimate the chances that a gap between two consecutive vehicles can be taken successfully. In contrast to previous models based on heuristic gap size rules, we present an approach which optimizes the integral risk of the situation using parametrized velocity ramps. It accounts for the risks from curves and all involved vehicles (front and rear on all paths) with a so-called survival analysis. For comparison, we also introduce a specially designed extension of the Intelligent Driver Model (IDM) for entering intersections.  We show in a quantitative statistical evaluation that the survival method provides advantages in terms of lower absolute risk (i.e., no crash happens) and better risk-utility tradeoff (i.e., making better use of appearing gaps). Furthermore, our approach generalizes to more complex situations with additional risk sources.

\end{abstract}

%
\IEEEpeerreviewmaketitle

\input{chapters/intro.tex}

\input{chapters/IDM.tex}
\input{chapters/ROPT.tex}
\input{chapters/experiments.tex}

\input{chapters/outlook.tex}

\section*{Acknowledgment}
\noindent This work has been partially supported by the European Unions Horizon 2020 project \textit{VI-DAS}, under the grant agreement number 690772. The authors would like to thank Fabian M\"uller for his support.




%



\bibliographystyle{IEEEtran}
\bibliography{bib.bib}

\end{document}

%% file: chapters/intro.tex
\vspace{0.2cm}
\section{Introduction}

Most Advanced Driver Assistance Systems (ADAS) enable semi-autonomous driving at low velocities in parking areas or at high velocities on highways, but not for moderate velocities in complex inner-city scenarios \cite{bengler2014}. They are separably developed, only reactive and not predictive and do not adapt to the driver's needs. 
Especially at intersections, the interplay of vehicles plays a dominant role and ADAS face a wide range of risk types (e.g. collision, curve, occlusion and traffic rules). 
This requires an ADAS that employs behavior prediction for risk estimation and subsequently plans safe behaviors. It has to generalize over scenarios and holistically incorporate different risk sources.

By optimizing integral risk and benefit factors over the prediction horizon employing the survival analysis \cite{eggert2014}, we compute trajectories consisting of velocity ramps on the ego path. 
The performance of ROPT is evaluated for merge-ins at unsignalized T-intersections. Merge-in support systems have sequential planning requirements due to curve taking and multiple other vehicles. 
We compare ROPT with an extended intersection version of the Intelligent Driver Model (IDM) \cite{treiber2000} using a statistical analysis. Particularly, we look at mean trends of gap number and gap size as well as of minimum back and front distance for varying parameter values of the models. 

The next Section \ref{sec:rel} gives an overview of related work focusing on the research community. The introduction of the Intersection IDM (IIDM) is divided into Section \ref{sec:curvidm} on longitudinal dynamics and Section \ref{sec:laneproj} on the lateral risk extension. We then explain more in detail ROPT's trajectory generation and optimization in Section \ref{traj_optim} and \ref{traj_eval}. Finally, Section \ref{sec:exp} shows the experiment setup together with results analysis and Section \ref{sec:outlook} a summary and discussion for future developments.

\subsection{Related Work}
\label{sec:rel}
Besides the automotive domain, numerous approaches for motion planning exist in robotics and physics. A survey of state of the art was conducted in \cite{paden2016}. 
Traffic simulators are categorized into microscopic and macroscopic models. 
While microscopic models, such as IDM, treat each car's dynamics seperately, macroscopic models look at traffic density, flow and average velocity with fluid dynamic equations \cite {francesco2015}. A subclass of microscopic models are cellular models, which discretize the time and space and thus work also well for larger road networks \cite{middleton1992}. 

The remaining planning methods are either using trajectory optimization or search algorithms. 
In \cite{ferguson2008}, velocity profiles of trapezoidal shapes are optimized with Model Predictive Control (MPC) and then evaluated against their proximity to dynamic obstacles, their smoothness and speed. 
Supplementary, \cite{naumann2017} evaluates besides the costs of the ego car, costs from the perspective of other cars to find a cooperative trajectory, which is applied to traffic scenes with priority orders. As a risk indicator, the Time-To-Collision (TTC) zone is used. Crossing of intersections can also be learned based on potential fields and the Levenberg-Marquardt method \cite{akagi2015}. Especially for T-intersections, \cite{orth2017} obtain personal critical gaps with Maximum Likelihood estimation and give out recommendations of safe gaps ahead.   

Standard search procedures find a path through static obstacle maps, but do not consider dynamic entities over time. For example \cite{chen2016} combine task planning with spatial exploration to detect free areas with circle-shaped spatial probability propagation. 
For this reason, Optimal Reciprocal Collision Avoidance (ORCA) is applied in \cite{berg2010} to search the velocity instead of position space in all directions and retrieve a collision-free velocity vector. 
Similarly, the authors of \cite{johnson2012} connect velocities along obstacle tangent points in the path-time space under acceleration bounds. 

%% file: chapters/IDM.tex
\section{Intersection IDM}
\label{sec:interidm}

\subsection{Basics and Curve Driving}
\label{sec:curvidm}
The IDM \cite{treiber2000} is a popular traffic model describing the dynamics of a leading and following vehicle pair driving along the same straight path. Its differential equation outputs a safe and efficient acceleration profile $\dot{v}_f$ for the follower

\begin{equation}
\dot{v}_f = \underbrace{a\left(1-\big(\frac{v_f}{v_c}\big)^\delta\right)}_{\mbox{\footnotesize free term}} -  \underbrace{a\left(\frac{d_0+v_fT}{d}+\frac{v_f(v_f-v_l)}{2d\sqrt{ab}}\right)^2\hspace{-0.15cm}}_{\mbox{\footnotesize interaction term}}. \\
\label{eq:idm}
\end{equation}

\noindent In general, the follower approaches the cruising velocity $v_c$ with the maximal acceleration $a$ and acceleration exponent $\delta$ in the free term. However, once another vehicle is in front at a distance $d$ with velocity $v_l$, the interaction term reduces $\dot{v}_f$ to reach a balance point defined by the minimal distance $d_0$ and time headway $T$. The contained reaction time can be set with the desired deceleration $b$.

Similar to \cite{liebner2013}, we additionally account for curves in the path by searching for the next maximal curvature $\kappa_{\mbox{\footnotesize max}}$ in a curve segment $[\kappa_{\mbox{\footnotesize start}}, \kappa_{\mbox{\footnotesize end}}]$. The resulting maximal lateral acceleration $a_y$ leads to an altered desired velocity
\begin{equation}
v_c = \sqrt{a_y/\kappa_{\mbox{\footnotesize max}}} \text{, if } \kappa_{\mbox{\footnotesize start}} > \kappa_{\mbox{\footnotesize th}} \text{ and } \kappa_{\mbox{\footnotesize end}} < \kappa_{\mbox{\footnotesize th}}.\
\label{eq:idmcurv}
\end{equation}
Hence, on sharp curve segments exceeding the threshold $\kappa_{\mbox{\footnotesize th}}$, the follower will converge to lower $v_f$ due to the free term. 

\subsection{Lane Projection}
\label{sec:laneproj}

Since the IDM is suitable for the simulation of longitudinal scenarios, it is often used in freeways. To improve the applicability, lane-change decisions have been modeled in \cite{kesting2007}. 
Here, the accelerations from the IDM  
are compared between the driver to the leader and the follower on the current as well as the adjacent lane. If the acceleration gain of the driver is higher than the loss of his followers, a lane change is executed. We extend this idea for the crossing of intersections.

First, we find the intersecting point between the current and other path and obtain the distance $d_I$ of the driver to the point. At segments close to the intersection, the position of the driver is projected and shifted along $d_I$ onto the other path. Then, we locate leading and following cars around this ``hypothetical'' driver. The IDM equations are taken for the driver relative to the leader and the follower relative to the driver to retrieve $\widetilde{a}_d$ and $\widetilde{a}_f$, respectively. For the case of the current path, we assume constant acceleration for the follower $a_f$ and a stop maneuver for the driver $a_d$. 
If the incentive criterion 
\begin{equation}
\underbrace{\widetilde{a}_d - a_d}_{\mbox{\footnotesize driver}} + p\underbrace{(\widetilde{a}_f - a_f)}_{\mbox{\footnotesize follower}}  > \Delta a_{\mbox{\footnotesize th}}  \\
\label{eq:iidm-go}
\end{equation}
is not fulfilled, the driver performs $a_d$. 
Otherwise the driver crosses the intersection by using $\widetilde{a}_d$. The politeness factor $p$ and threshold $\Delta a_{\mbox{\footnotesize th}}$ modify the needed advantage at which the the driver will pass in front of the follower. 

With the incentive criterion alone, it is possible that the follower crashes into the driver for small $p$. 
We therefore check simultaneously the safety criterion  
\begin{equation}
\widetilde{a}_f \ge b_{s} \\
\label{eq:iidm}
\end{equation}
with a safe deceleration $b_s$. The resulting method is called the IIDM.

%% file: chapters/ROPT.tex
\section{Risk Optimization Method}
\label{sec:ropt}
In each time step of the simulation, ROPT receives information about the current state of the environment. 
This includes the latest measured position and velocity of all traffic participants as well as their associated future paths extracted from map data. The goal of ROPT is to predict an optimal velocity profile for the ego car along its path. The computation consists of three basic steps. Initially, a trajectory for each  other car is extrapolated over the prediction horizon.\footnote{ROPT uses a constant velocity assumption. However, it is also possible to use other methods, e.g. constant acceleration, or variants.} Second, a set of potential ego trajectories are created.  
Third, one of the created ego trajectories is selected based on its integral risk (caused by curve and collision), utility (distance travelled) and comfort (strength and frequency of velocity change). Since we use an optimization algorithm for the trajectory generation, steps two and three are heavily intertwined. 
\subsection{Trajectory Optimization}
\label{traj_optim}
For scenarios with only one risk source (i.e., car following, curve driving and driving straight on an intersection), basic trajectory sampling methods are usually sufficient. Accordingly in previous research, the Foresighted Driver Model (FDM) \cite{damerow2015fdm} was developed. It samples via gradient descent an acceleration and deceleration profile and balances risk with utility to find correct ego behaviors.

Merge-ins at intersections are however more complex scenarios in which the planned velocity profile has to obey maximum curve speed and match the speed of the traffic flow. The FDM would only converge to the local minimum from the curve risk. Here, it is necessary to generate a velocity profile that considers multiple spatio-temporal risk sources. 
In a different approach, we incrementally constructed velocity paths through predictive risk maps with Rapidly-exploring Random Trees (RRT) \cite{damerow2015}. It was proven to work in intersection scenarios with multiple other cars, but the step-wise evaluation makes it harder to constrain the convergence of the solution to specific simple trajectories.

\begin{figure}[t!]
\centering
\vspace{0.11cm}
\resizebox{0.48\textwidth}{!}{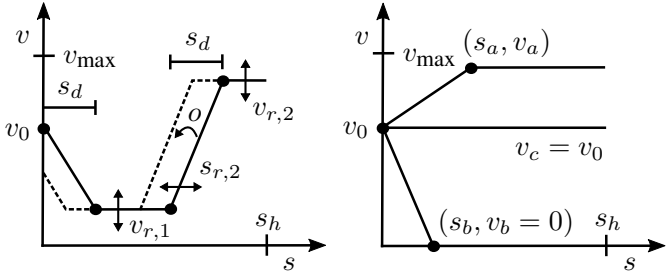}
\caption{Left: Optimization of velocity ramps for predicted times. Right: Fixed standard trajectories.} %
\vspace{0.15cm}
\label{fig:ramp}
\end{figure}

ROPT therefore optimizes parametrized velocity profiles consisting of two consecutive acceleration and deceleration ramps (see left-hand side of Figure \ref{fig:ramp}). A double-ramp profile is described by the end velocities of the first and second ramp $v_{r,1}$ and $v_{r,2}$ plus the start time of the second ramp $s_{r,2}$. Each ramp has a fixed duration $s_d\hspace{-0.05cm}=\hspace{-0.05cm}\unit[2.5]{s}$, whereas the first ramp starts with the current velocity $v_0$.

We use the Nelder-Mead opimization algorithm \cite{nelder1965simplex}, a downhill simplex method that does not require gradient information. In each optimization iteration, the simple double-ramp velocity profile is converted into a trajectory, which in turn is evaluated for risk, utility and comfort (see Section \ref{traj_eval}). Trajectories that violate constraints on velocity or accleration are penalized. Since the Nelder-Mead optimization is a local search, it depends on the initial value of the parameters. Hence, ROPT simultaneously optimizes $k$ trajectories, starting from different initial values $v_{r,1}^j = v_{r,2}^j = v_r^j$ with $j\in{1,\dots,k}$ and $v_r^j$ evenly spaced in $[\unit[2]{m/s},v_\text{max}]$.
If trajectory $j$ was chosen in the previous time step, the optimization continues with its previous parameters. Additionally, it shifts the beginnings and ends of all ramps by an offset $o$ that corresponds to the total time that the trajectory has been selected.\footnote{If the trajectory has been active for a duration equal to the ramp length ($o=s_r$), we insert a new ramp after ramp two if $s_2<s_r$ or before ramp two otherwise.} In other words, if a planned trajectory is executed for multiple time steps, it is time-shifted and fine tuned in each step.

In addition to the optimized trajectries, ROPT samples three simple trajectories (see right-hand side of Figure \ref{fig:ramp}): A constant velocity trajectory $v_c = v_0$, a stopping trajectory with the end point $(s_b, v_b = 0)$ and an acceleration trajectory to the set speed $v_a$ at the predicted time $s_a$. 

\vspace{0.15cm}

\subsection{Trajectory Evaluation}
\label{traj_eval}
\subsubsection{Risk Prediction}

\begin{figure}[t!]
  \centering
  \vspace{-0.2cm}
  \resizebox{1.0\linewidth}{!}{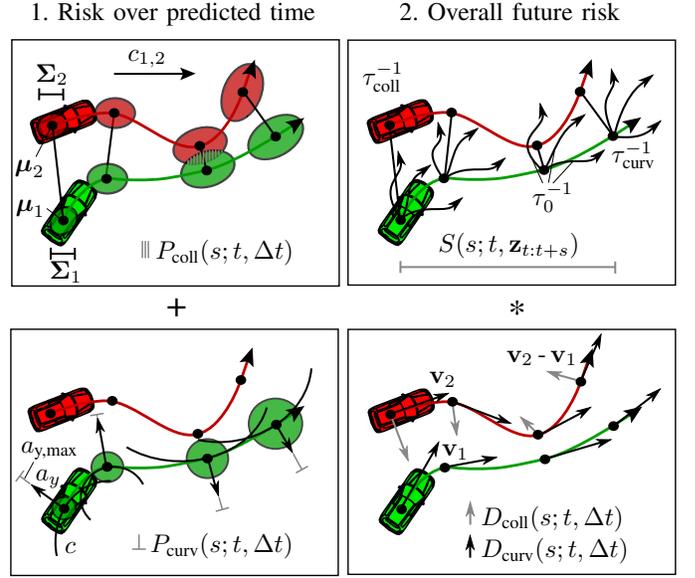}
  \caption{Left: Gaussian method for collision and curve probability. Right: Survival analysis with corresponding event rates and damage terms.}
  \label{fig:combgs}
\end{figure}

An indicator for risk is the probability function $P_{\text{crit}}(s;t,\Delta t)$ that a critical event will happen from the current time $t$ around a future time $t+s$ during an interval of size $\Delta t$. A compact risk measure $R(t)$ then comprises the entire accumulated future risk contained in $P_{\text{crit}}(s;t,\Delta t)$, $s\in[0,\infty]$. By also weighting $P_{\text{crit}}(s;t,\Delta t)$ with the predicted damage of the event $D_{\text{crit}}(s;t,\Delta t)$, 
we obtain risk as the expected future severity. ROPT combines the Gaussian method \cite{garmier2009} for the estimation of critical event probabilities with the survival analysis \cite{eggert2014} to gain the overall risk.
In the following, we consider the situation evolution of an ego car (green) encountering another car (red) indexed with $i\hspace{-0.05cm}=\hspace{-0.05cm}1,2$ as depicted in Figure \ref{fig:combgs}. 
The car's future trajectories are predicted as in Section \ref{traj_optim}. 

It cannot be assumed that the cars follow exactly the predicted trajectory. In reality, they undergo variations in speed and heading. Consequently, we model their respective spatial position with a normal distribution $f_{i}$.
Since we predict the vehicles to drive along predefined paths, we furthermore define that the longitudinal uncertainty along the path is higher than the lateral uncertainty. 
In this way, we obtain 2D ellipses that are specified by an uncertainty matrix $\mathbf{\Sigma}^\text{'}_{i}$ around the mean position vector $\boldsymbol{\mu}_{i}$ with
\begin{equation}
\boldsymbol{\mu}_{i} = 
  \begin{bmatrix}
    {\mu}_{i,x} \\
    {\mu}_{i,y} \\
  \end{bmatrix}, \quad
\mathbf{\Sigma}^\text{'}_{i} = 
  \begin{bmatrix}
    \sigma^2_{i,\text{lon}} & 0 \\
    0 & \sigma^2_{i,\text{lat}} \\
  \end{bmatrix} .
\label{eq:musigma}
\end{equation}

\noindent A collision occurs if both cars coincide at the same position, which is analog to $f_{\text{coll}}=f_1f_2$. 
To retrieve the product of two Gaussian functions, the uncertainties $\mathbf{\Sigma}^\text{'}_{i}$ have to be transformed into the same global coordinate system $x,y$ according to
\begin{equation}
\mathbf{\Sigma}_{i} = \mathbf{R} \mathbf{\Sigma}^\text{'}_{i} \mathbf{R}^T \ \text{and} \
\mathbf{R} = 
  \begin{bmatrix}
    \cos \alpha_{i} & -\sin \alpha_{i} \\
    \sin \alpha_{i} & \cos \alpha_{i} \\
  \end{bmatrix}. 
\label{eq:rotsigma}
\end{equation}
The collision probability is eventually given by spatially integrating $f_{\text{coll}}$ over all positions 
\vspace{0.2cm}
\begin{align}
P_{\text{coll}}(s;t,\Delta t) = |2\pi&(\mathbf{\Sigma}_{1}+\mathbf{\Sigma}_{2})|^{-\frac{1}{2}} * \nonumber \\ \exp\{-\frac{1}{2}
(\boldsymbol{\mu}_{2}-\boldsymbol{\mu}_{1})^T(\mathbf{\Sigma}_{1}&+\mathbf{\Sigma}_{2})^{-1}
(\boldsymbol{\mu}_{2}-\boldsymbol{\mu}_{1})\}.   
\label{eq:prodgauss}
\end{align}

Due to temporal uncertainty, the deviation of the real from the predicted trajectory differs with increasing prediction times. We extrapolate the kinematics of the current state to achieve trajectories. After a prediction step of size $\Delta s$, their longitudinal position on the path $l_{i}$
is shifted by $\Delta l_{i}$ and we get 
\begin{equation}
l_{i}(s+1) = l_{i}(s) + \Delta l_{i} = l_{i}(s) + v_{i}(s) \Delta s
\label{eq:diststep}
\end{equation}
with velocities $v_{i}$.
Knowing this, the growth of spatial uncertainty can be derived from a velocity uncertainty factor $c_{i}$ using
\begin{equation}
\sigma_{i,v}(s+1) := \sigma_{i,v}(s) + c_{i} v_{i}(s) \Delta s.  
\label{eq:sigmavstep}
\end{equation}

The probability for the ego car to drive off at sharp curves is formulated similarly. Here, we assume 1D circles with uncertainty $\sigma_1$ and look at the lateral acceleration 
\begin{equation}
a_{y}(s) = \sqrt{\rule{0pt}{0.3cm}\kappa v_{\text{1}}},
\end{equation}
which is influenced by the curvature of the road $\kappa$.
We then compare $a_y$ with its maximal possible value $a_{\text{y,max}}$ from vehicle dynamics constraints to retrieve
\vspace{0.1cm}
\begin{equation}
P_{\text{curv}}(s;t,\Delta t) = \frac{1}{\sqrt{2\pi\sigma^2_{1}}} \, \exp\left\{-\frac{\text{max}(a_{\text{y,max}}-|a_y|,0)^2}{2\sigma_{1}^2}\right\}. \nonumber
\end{equation}
\vspace{0.03cm}

\noindent If $|a_y|$ approaches $a_{\text{y,max}}$, the probability $P_{\text{curv}}(s;t,\Delta t)$ will thus increase.
\vspace{0.09cm}

Next in the survival analysis, accident occurrences are modeled as a thresholding process based on Poisson-like event probabilities. 
A Poisson process is defined by a situation state-dependent total event rate $\tau^{-1}(\textbf{z}_{t:t+s})$, which characterizes the mean time between events and consists of a critical event rate $\tau^{-1}_{\text{crit}}$ and a constant escape rate $\tau^{-1}_0$ (comprising behavioral options that mitigate critical events). For $\tau^{-1}_{\text{crit}}$, we consider collision risks represented by the single event rate $\tau_{\text{coll}}^{-1}$ and the risk of losing control in curves $\tau^{-1}_{\text{curv}}$ which leads to
\begin{equation}
\tau^{-1}(\textbf{z}_{t:t+s})=\tau^{-1}_0 + \tau^{-1}_{\text{crit}}=\tau^{-1}_0+\tau^{-1}_{\text{coll}}+\tau^{-1}_{\text{curv}}  
\end{equation}
with 
\begin{equation}
\tau_{\text{crit}}^{-1}(\textbf{z}_{t:t+s}) = P_{\text{crit}}(s;t,\Delta t) / \Delta t,
\end{equation}
\begin{equation}
P_{\text{crit}}(s;t,\Delta t) = P_{\text{coll}}(s;t,\Delta t) + P_{\text{curv}}(s;t,\Delta t).
\end{equation}
%

\noindent The survival function indicates the probability that the vehicle will not be engaged in an event like an accident from $t$ until $t+s$ in compliance with
\begin{equation}
S(s;t,\textbf{z}_{t:t+s})=\exp\{-\int_0^s \tau^{-1}(\textbf{z}_{t:t+s'}) \, ds'\}.
\end{equation}

It has been empirically shown that human injury compared to the kinetic energy of the accident (proportional to velocity vectors $\mathbf{v}_{i}$) has the behavior of a logistic function \cite{chen2010}. We postulate that the damage of the involved cars has the same qualitative relationship. Car-to-car collision and curve control loss damage is given by
\begin{equation}
D_{\text{coll}}(s;t,\Delta t) = \frac{D_{\footnotesize \mbox{max,coll}}}{1+\mbox{exp}\{ k_{\footnotesize \mbox{coll}} (\| \mathbf{v}_2 - \mathbf{v}_1 \| - \beta_{\footnotesize \mbox{coll}})  \}}, \\
\label{eq:damagecoll}
\end{equation}
\begin{equation}
D_{\text{curv}}(s;t,\Delta t) = \frac{D_{\footnotesize \mbox{max,curv}}}{1+\mbox{exp}\{ k_{\footnotesize \mbox{curv}}(\| \mathbf{v}_1 \| - \beta_{\footnotesize \mbox{curv}})  \}}, \\
\label{eq:damagecurv}
\end{equation}
\vspace{0.02cm}
\noindent where the parameter $k$ is the damage increase factor, $\beta$ the damage midpoint and $D_{\footnotesize \mbox{max}}$ the maximal damage.
As a result, we acquire the overall risk engaging in a future critical event by temporally integrating the term of probabilities, damages and survival function 
\begin{equation}
R(t) = \int_0^{\infty} (\tau_{\text{coll}}^{-1}D_{\text{coll}}+\tau_{\text{curv}}^{-1}D_{\text{curv}})S \,ds.
\end{equation}

\subsubsection{Utility and Comfort Prediction}
\noindent A driver tries to minimize the risk, but maximize his benefit as well. In ROPT, the considered benefit consists of the needed time to arrive at the goal and the comfort of the travel. The former is 
defined by the velocity course of the ego car $v_1$ and the latter takes the acceleration and jerk profile $a_1$ and $j_1$ into account.\footnote{The double ramp parameters have limited influence in $j_1$. Here, polynoms as velocity profiles can be beneficial due to their property of continuity.} We weight the components with driver-specific constants $b^t$, $b^c$ and $b^j$ and compute the integral future benefit with
\begin{equation}
B(t)=\int_0^{\infty} (b^t |v_1|-b^c |a_1|- b^j |j_1|) S \,ds.
\end{equation} 
For higher $s$, we also consider the survival function $S$ in the evaluation. In this way, high-risk situations result into lower $B(t)$.
ROPT finally evaluates for each generated trajectory the cost function
\begin{equation}
C(t) = R(t) - B(t) \\
\label{eq:cost}
\end{equation}
and executes the trajectory with the lowest $C(t)$. For comparability, $R(t)$ and $B(t)$ have to be transformed into the same unit. Since the severity factors in $R(t)$ require a monetarization, we use \euro \hspace{0.04cm} for $C(t)$.

%% file: img/vel_sampling4.pdf_tex
\begingroup%
  \makeatletter%
  \providecommand\color[2][]{%
    \errmessage{(Inkscape) Color is used for the text in Inkscape, but the package 'color.sty' is not loaded}%
    \renewcommand\color[2][]{}%
  }%
  \providecommand\transparent[1]{%
    \errmessage{(Inkscape) Transparency is used (non-zero) for the text in Inkscape, but the package 'transparent.sty' is not loaded}%
    \renewcommand\transparent[1]{}%
  }%
  \providecommand\rotatebox[2]{#2}%
  \ifx\svgwidth\undefined%
    \setlength{\unitlength}{230.260586bp}%
    \ifx\svgscale\undefined%
      \relax%
    \else%
      \setlength{\unitlength}{\unitlength * \real{\svgscale}}%
    \fi%
  \else%
    \setlength{\unitlength}{\svgwidth}%
  \fi%
  \global\let\svgwidth\undefined%
  \global\let\svgscale\undefined%
  \makeatother%
  \begin{picture}(1,0.41091186)%
    \put(0,0){\includegraphics[width=\unitlength,page=1]{vel_sampling4.pdf}}%
    \put(0.26615162,0.34521089){\color[rgb]{0,0,0}\makebox(0,0)[lb]{\smash{$s_d$}}}%
    \put(0.36915162,0.23821089){\color[rgb]{0,0,0}\makebox(0,0)[lb]{\smash{$v_{r,2}$}}}%
    \put(0.28792341,0.1484678){\color[rgb]{0,0,0}\makebox(0,0)[lb]{\smash{$s_{r,2}$}}}%
    \put(0.41530864,0.00113325){\color[rgb]{0,0,0}\makebox(0,0)[lb]{\smash{$s$}}}%
    \put(0.1809166,0.05888455){\color[rgb]{0,0,0}\makebox(0,0)[lb]{\smash{$v_{r,1}$}}}%
    \put(0.270469308,0.22980629){\color[rgb]{0,0,0}\makebox(0,0)[lb]{\smash{$o$}}}%
    \put(0.37297534,0.069339){\color[rgb]{0,0,0}\makebox(0,0)[lb]{\smash{$s_h$}}}%
    \put(0.071101465,0.27120211){\color[rgb]{0,0,0}\makebox(0,0)[lb]{\smash{$s_d$}}}%
    \put(-0.009101465,0.20720211){\color[rgb]{0,0,0}\makebox(0,0)[lb]{\smash{$v_{0}$}}}%
    \put(0.00739433,0.34853258){\color[rgb]{0,0,0}\makebox(0,0)[lb]{\smash{$v$}}}%
    \put(0.08043146,0.31808405){\color[rgb]{0,0,0}\makebox(0,0)[lb]{\smash{$v_{\mbox{\footnotesize max}}$}}}%
    \put(0.50730416,0.20977035){\color[rgb]{0,0,0}\makebox(0,0)[lb]{\smash{$v_0$}}}%
    \put(0.68956356,0.33848039){\color[rgb]{0,0,0}\makebox(0,0)[lb]{\smash{$(s_a, v_a)$}}}%
    \put(0.77054989,0.17690568){\color[rgb]{0,0,0}\makebox(0,0)[lb]{\smash{$v_c=v_0$}}}%
    \put(0.65226353,0.06342927){\color[rgb]{0,0,0}\makebox(0,0)[lb]{\smash{$(s_b, v_b=0)$}}}%
    \put(0.93552848,0.00141339){\color[rgb]{0,0,0}\makebox(0,0)[lb]{\smash{$s$}}}%
    \put(0.52585565,0.34788165){\color[rgb]{0,0,0}\makebox(0,0)[lb]{\smash{$v$}}}%
    \put(0.59889271,0.32048405){\color[rgb]{0,0,0}\makebox(0,0)[lb]{\smash{$v_{\mbox{\footnotesize max}}$}}}%
    \put(0.89143666,0.06868807){\color[rgb]{0,0,0}\makebox(0,0)[lb]{\smash{$s_h$}}}%
  \end{picture}%
\endgroup%

%% file: img/gauss_plus_survival_final8.pdf_tex
\begingroup%
  \makeatletter%
  \providecommand\color[2][]{%
    \errmessage{(Inkscape) Color is used for the text in Inkscape, but the package 'color.sty' is not loaded}%
    \renewcommand\color[2][]{}%
  }%
  \providecommand\transparent[1]{%
    \errmessage{(Inkscape) Transparency is used (non-zero) for the text in Inkscape, but the package 'transparent.sty' is not loaded}%
    \renewcommand\transparent[1]{}%
  }%
  \providecommand\rotatebox[2]{#2}%
  \ifx\svgwidth\undefined%
    \setlength{\unitlength}{261.21622639bp}%
    \ifx\svgscale\undefined%
      \relax%
    \else%
      \setlength{\unitlength}{\unitlength * \real{\svgscale}}%
    \fi%
  \else%
    \setlength{\unitlength}{\svgwidth}%
  \fi%
  \global\let\svgwidth\undefined%
  \global\let\svgscale\undefined%
  \makeatother%
  \begin{picture}(1,0.90350505)%
    \put(0,0){\includegraphics[width=\unitlength,page=1]{gauss_plus_survival_final8.pdf}}%
    \put(0.78421869,0.56701672){\color[rgb]{0,0,0}\makebox(0,0)[lb]{\smash{$\tau_{0}^{-1}$}}}%
    \put(0.643248258,0.49607882){\color[rgb]{0,0,0}\makebox(0,0)[lb]{\smash{$S(s;t, \mathbf{z}_{t:t+s})$}}}%
    \put(0.90315771,0.63127401){\color[rgb]{0,0,0}\makebox(0,0)[lb]{\smash{$\tau_{\mbox{\scriptsize curv}}^{-1}$}}}%
    \put(0.0301836,0.84460878){\color[rgb]{0,0,0}\makebox(0,0)[lb]{\smash{1. Risk over predicted time}}}%
    \put(0.585035929,0.84484169){\color[rgb]{0,0,0}\makebox(0,0)[lb]{\smash{2. Overall future risk}}}%
    \put(0.52883674,0.74799655){\color[rgb]{0,0,0}\makebox(0,0)[lb]{\smash{$\tau_{\mbox{\scriptsize coll}}^{-1}$}}}%
    \put(0.03877837,0.75280349){\color[rgb]{0,0,0}\makebox(0,0)[lb]{\smash{$\mathbf{\Sigma}_{2}$}}}%
    \put(0.17699722,0.78417778){\color[rgb]{0,0,0}\makebox(0,0)[lb]{\smash{$c_{1,2}$}}}%
    \put(0.21880254,0.48622203){\color[rgb]{0,0,0}\makebox(0,0)[lb]{\smash{$P_{\mbox{\scriptsize coll}}(s;t,\scriptsize \Delta t)$}}}%
    \put(0.05958084,0.45571771){\color[rgb]{0,0,0}\makebox(0,0)[lb]{\smash{$\mathbf{\Sigma}_{1}$}}}%
    \put(0.00886217,0.55853288){\color[rgb]{0,0,0}\makebox(0,0)[lb]{\smash{$\boldsymbol{\mu}_1$}}}%
    \put(0.00744564,0.61774399){\color[rgb]{0,0,0}\makebox(0,0)[lb]{\smash{$\boldsymbol{\mu}_2$}}}%
    \put(0.23519209,0.39922726){\color[rgb]{0,0,0}\makebox(0,0)[lb]{\smash{$\mbox{\Large +}$}}}%
    \put(0.74796313,0.38827531){\color[rgb]{0,0,0}\makebox(0,0)[lb]{\smash{$\mbox{\Large *}$}}}%
    \put(0.03856394,0.15272887){\color[rgb]{0,0,0}\makebox(0,0)[lb]{\smash{$a_y$}}}%
    \put(0.08249423,0.04353325){\color[rgb]{0,0,0}\makebox(0,0)[lb]{\smash{$c$}}}%
    \put(0.21087588,0.04577478){\color[rgb]{0,0,0}\makebox(0,0)[lb]{\smash{$P_{\mbox{\scriptsize curv}}(s;t,\scriptsize \Delta t)$}}}%
    \put(0.01657358,0.19571445){\color[rgb]{0,0,0}\makebox(0,0)[lb]{\smash{$a_{\mbox{\scriptsize y,max}}$}}}%
    \put(0.64844789,0.1905348){\color[rgb]{0,0,0}\makebox(0,0)[lb]{\smash{$\mathbf{v}_1$}}}%
    \put(0.7070852,0.0864691){\color[rgb]{0,0,0}\makebox(0,0)[lb]{\smash{$D_{\mbox{\scriptsize coll}}(s;t,\scriptsize \Delta t)$}}}%
    \put(0.7070852,0.0374691){\color[rgb]{0,0,0}\makebox(0,0)[lb]{\smash{$D_{\mbox{\scriptsize curv}}(s;t,\scriptsize \Delta t)$}}}%
    \put(0.62949885,0.29969836){\color[rgb]{0,0,0}\makebox(0,0)[lb]{\smash{$\mathbf{v}_2$}}}%
    \put(0.74557306,0.33150554){\color[rgb]{0,0,0}\makebox(0,0)[lb]{\smash{$\mathbf{v}_2$\hspace{0.06cm}-\hspace{0.06cm}$\mathbf{v}_1$}}}%
  \end{picture}%
\endgroup%

%% file: chapters/experiments.tex
\section{Experiments}
\label{sec:exp}
\subsection{Setup}
\begin{figure}
  \centering
  \resizebox{1.0\linewidth}{!}{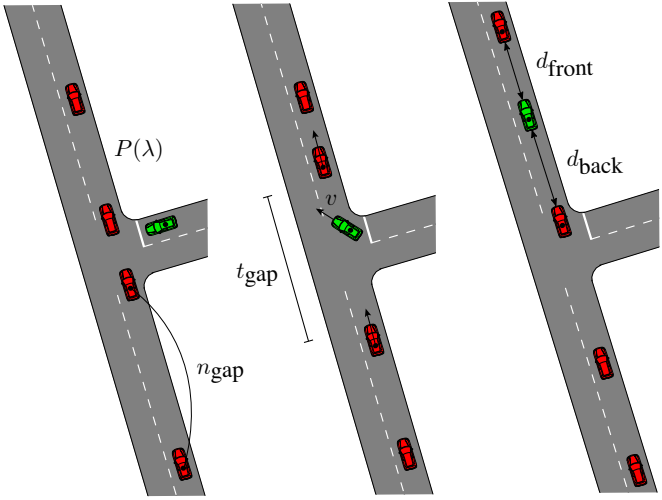}
  \caption{Time lapse of T-intersection merge-in scenario and relevant evaluation variables.}
  \label{fig:scenario}
\end{figure}

In our simulation, IIDM and ROPT face the complex task of planning succesful merge-ins at T-intersections.  
Figure \ref{fig:scenario} illustrates the ego car waiting at the stop line, while other cars pass from left to right. We model the traffic as a Poisson distribution $P(\lambda)$ with different intervals $\lambda$ and count the number of missed gaps $n_\text{gap}$.\footnote{For continuous $\lambda$, we add a uniform random noise offset in the range of $[\unit[-0.5]{s},\unit[0.5]{s}]$.} If the gap size $t_\text{gap}$ is large enough, the ego car needs to increase $v$ until the maximum curve velocity and subsequently accelerate even further to match the velocity of the traffic flow.  
The angle of the right turn with $\unit[90]{^\circ}$ and the constant traffic speed with $v_f=\unit[10]{m/s}$ are chosen in a way that makes single-ramp trajectories only feasible for very large gaps of more than $\unit[100]{m}$.

Besides systematically changing $\lambda$ which results in different traffic densities, we modify the politeness factor $p$ of IIDM and the travel benefit $b^t$ for ROPT to control the driving behavior.    
For every model and traffic setting, we then run several simulations (ca. 200) to ensure statistical significance. 
Once the ego car advances past the stop line, the merge-in procedure has started and the encountered distances to the front $d_\text{front}$ and rear vehicle $d_\text{back}$ are recorded. 
In the evaluation, we look at the mean minimal distances $\bar{d}_\text{back,min}$ and $\bar{d}_\text{front,min}$, which indicate risk, as well as the mean utility indicators $\bar{n}_\text{gap}$ and $\bar{t}_\text{gap}$.\footnote{Ideal environment conditions are assumed, in which position and velocity of the involved cars are synchronized and known without sensor errors.} 

\subsection{Results}
\begin{figure}[t!]
  \centering
  \vspace{-0.46cm}
  \resizebox{1.0\linewidth}{!}{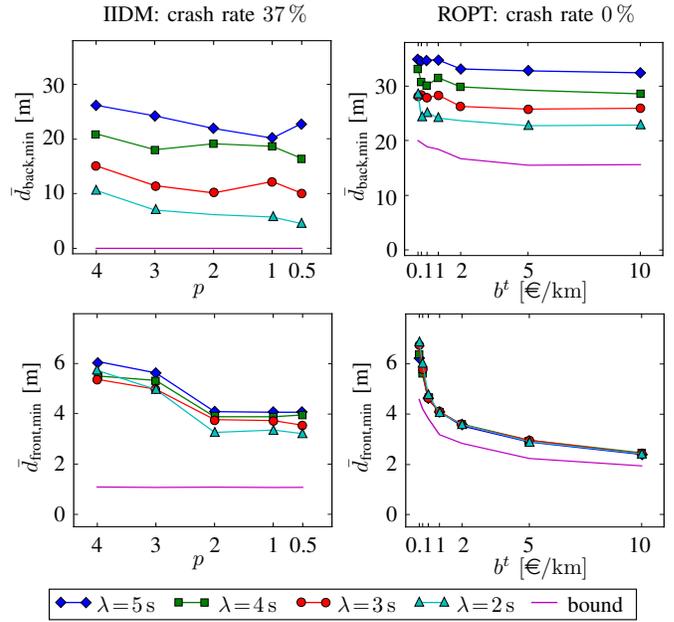}
  \caption{Comparison of IIDM and ROPT with risk indicators.} 
  \label{fig:comprisk}
\end{figure}

Figure \ref{fig:comprisk} plots the distances $\bar{d}_\text{back,min}$ and $\bar{d}_\text{front,min}$ for IIDM with $p \hspace{-0.06cm} = \hspace{-0.06cm} [0.5,4]$ and ROPT with $b^t\hspace{-0.06cm}=\hspace{-0.06cm}[\unit[0.1]{\text{\euro}/km},\unit[10]{\text{\euro}/km}]$  
for different $\lambda\hspace{-0.06cm}=\hspace{-0.06cm}[\unit[2]{s},\unit[5]{s}]$. 
As expected, $\bar{d}_\text{back,min}$ decreases with lower $\lambda$ for both models. Likewise, it declines with lower $p$ or higher $b^t$. The ego car enters smaller gaps, where the rear car approaches more closely due to its excess speed.
More importantly, the lower bound for $\bar{d}_\text{back,min}$ over all runs (purple line) is always $0$ in the case of the IIDM. Merge-ins cannot be performed properly with accidents happening in $\unit[37]{\%}$ of the runs.\footnote{A crash or near-crash case is assumed, if the distance between two cars is less than $\unit[1]{m}$.}
The reason can be identified in the simulation: At time step $t_{\text{start}}$, the IIDM decides to merge in based on the incentive criterion of Equation (\ref{eq:iidm-go}). However due to the restricted curve speed, 
it cannot accelerate quickly enough in front of the follower. At some later time step $t_{\text{end}}$, the criterion becomes invalid and a brake is executed. 
If the resulting stopping position is in the way of the crossing traffic, an accident occurs. 
In contrast, the lower bound of ROPT stays above $\unit[15]{m}$. ROPT does not expect any deceleration, but assumes constant velocity for the follower and yields consistently safe behavior.

Regarding $\bar{d}_\text{front,min}$, only runs with safe mergers are listed for IIDM (otherwise there is no notion of a back car). While $\bar{d}_\text{front,min}$ is approximately the same for IIDM and ROPT with different $\lambda$, it also  
descends with smaller $p$ or bigger $b^t$. 
In ROPT, bigger $b^t$ lead to less weight on risk and earlier merge-in begins. With IIDM, $d_\text{front}$ is usually constant. Equation (\ref{eq:iidm-go}) does not consider the front car. Nevertheless, due to unsuccessful merge-in attempts at previous gaps, it is possible that the ego car has already advanced into the intersection.

\begin{figure}[t!]
  \centering
  \vspace{-0.3cm}
  \resizebox{1.0\linewidth}{!}{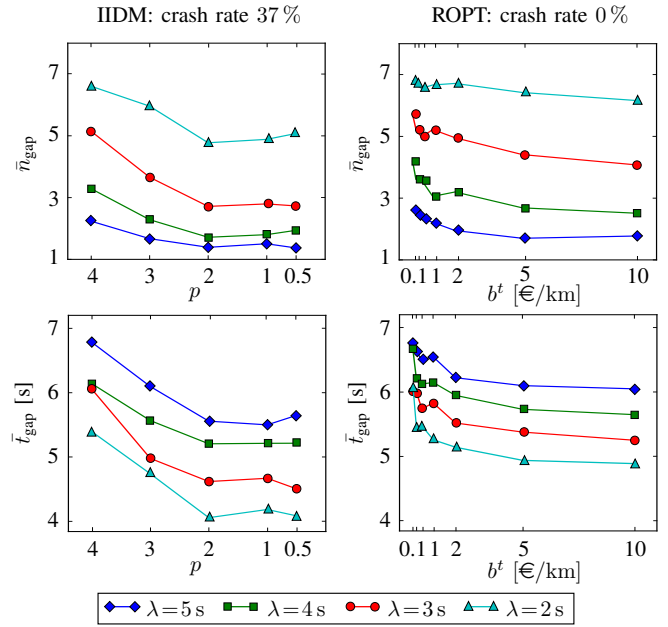}
  \caption{Comparison of IIDM and ROPT with utility indicators.} 
  \label{fig:computil}
\end{figure}

Figure \ref{fig:computil} pictures $\bar{n}_\text{gap}$ and $\bar{t}_\text{gap}$ with the same parameter variations.
Both models choose earlier gaps (less $\bar{n}_\text{gap}$) based on willingness of driving into lower $\bar{t}_\text{gap}$ with decreasing $p$ or increasing $b^t$. In general, IIDM has sharper slopes, because of its risk-proneness.    
For dense traffic settings (small $\lambda$), the models need to wait longer with higher $\bar{n}_\text{gap}$ and the taken gap has lower $\bar{t}_\text{gap}$. In total, $\bar{t}_\text{gap}$ lies in between $\unit[4 - 7]{s}$ which are realistic critical gaps \cite{orth2017}.

\subsubsection{Predictive IIDM}

We saw that IIDM is unable to cope with merge-in scenarios. 
It implicitly predicts the behavior of the driver and leading vehicle, but lacks a proper prediction of the follower. For this reason, we extend the IIDM further. In the spirit of ROPT, we explicitly extrapolate trajectories of other cars with constant velocity and the ego trajectory using Equations (\ref{eq:idm}) and (\ref{eq:idmcurv}). At every predicted step, the criterions (\ref{eq:iidm-go}) and (\ref{eq:iidm}) are evaluated. Only if both hold true for the complete horizon, a merge-in is started.
The outputs of the predictive IIDM are shown in Figure \ref{fig:pred-idm}. The lower bound for $\bar{d}_\text{back,min}$ approaches $\unit[14]{m}$ for $p\hspace{-0.04cm}\rightarrow\hspace{-0.04cm}0.5$ , because $b_{safe}$ is set to $\unit[-2]{m/s^2}$ in the safety criterion. The courses of $\bar{d}_\text{front,min}$ are constant $\unit[7]{m}$ for different $\lambda$. Independent of $p$ and $\lambda$, a safe behavior is achieved. Additionally, $\bar{n}_\text{gap}$ and $\bar{t}_\text{gap}$ are shifted to higher values and similar to ROPT.  



\subsubsection{Discussion}
ROPT distributes the available $t_\text{gap}$ more to $d_\text{front}$ than $d_\text{back}$, while keeping the absolute risk low (i.e . no crash happens) for varying $b^t$ and $\lambda$. As opposed to this, IIDM is not able to cap the risk with its heuristics. The predictive IIDM generates safe behavior, but different $p$ can only change its influence to $d_\text{back}$. This is undesirable, because in merge-ins the behavior of front cars is better manageable than back cars. Moreover, neither versions of IIDM can derive a continous acceleration course. They only switch between braking or following the front car. ROPT inherently possesses an explicit risk-utility tradeoff (i.e., making better use of $t_\text{gap}$). 

For intersections of arbitrary geometry and traffic constellations, ROPT works out of the box due to its instantaneous adaptation using a full holistic predictive risk model. For instance, a crossing of X-intersections could be also handled in the predictive IIDM with additional go/no-go decisions. However, lateral risks are only considered shortly before the intersection and not afterwards. In complex interactions between the ego and multiple other cars, the combination of heuristics will potentially lead to unsafe behavior.

%% file: img/gap_scenario_final5.pdf_tex
\begingroup%
  \makeatletter%
  \providecommand\color[2][]{%
    \errmessage{(Inkscape) Color is used for the text in Inkscape, but the package 'color.sty' is not loaded}%
    \renewcommand\color[2][]{}%
  }%
  \providecommand\transparent[1]{%
    \errmessage{(Inkscape) Transparency is used (non-zero) for the text in Inkscape, but the package 'transparent.sty' is not loaded}%
    \renewcommand\transparent[1]{}%
  }%
  \providecommand\rotatebox[2]{#2}%
  \ifx\svgwidth\undefined%
    \setlength{\unitlength}{285.46276065bp}%
    \ifx\svgscale\undefined%
      \relax%
    \else%
      \setlength{\unitlength}{\unitlength * \real{\svgscale}}%
    \fi%
  \else%
    \setlength{\unitlength}{\svgwidth}%
  \fi%
  \global\let\svgwidth\undefined%
  \global\let\svgscale\undefined%
  \makeatother%
  \begin{picture}(1,0.72561151)%
    \put(0.67427571,0.04044561){\color[rgb]{0,0,0}\rotatebox{1.27283189}{\makebox(0,0)[lb]{\smash{}}}}%
    \put(0,0){\includegraphics[width=\unitlength,page=1]{gap_scenario_final5.pdf}}%
    \put(0.353353528,0.32694438){\color[rgb]{0,0,0}\rotatebox{1.27283189}{\makebox(0,0)[lb]{\smash{$t_{\mbox{gap}}$}}}}%
    \put(0.4871264495,0.43681006){\color[rgb]{0,0,0}\rotatebox{1.27283303}{\makebox(0,0)[lb]{\smash{$v$}}}}%
    \put(0.80316169,0.64228037){\color[rgb]{0,0,0}\rotatebox{1.27283303}{\makebox(0,0)[lb]{\smash{$d_{\mbox{front}}$}}}}%
    \put(0.84956159,0.49562199){\color[rgb]{0,0,0}\rotatebox{1.27283303}{\makebox(0,0)[lb]{\smash{$d_{\mbox{back}}$}}}}%
    \put(0.1700126,0.5111931){\color[rgb]{0,0,0}\makebox(0,0)[lb]{\smash{$P(\lambda)$}}}%
    \put(0.29394716,0.1827962){\color[rgb]{0,0,0}\rotatebox{1.27283189}{\makebox(0,0)[lb]{\smash{$n_{\mbox{gap}}$}}}}%
  \end{picture}%
\endgroup%

%% file: img/risk_ind_ROPT_IIDM.pdf_tex
\begingroup%
  \makeatletter%
  \providecommand\color[2][]{%
    \errmessage{(Inkscape) Color is used for the text in Inkscape, but the package 'color.sty' is not loaded}%
    \renewcommand\color[2][]{}%
  }%
  \providecommand\transparent[1]{%
    \errmessage{(Inkscape) Transparency is used (non-zero) for the text in Inkscape, but the package 'transparent.sty' is not loaded}%
    \renewcommand\transparent[1]{}%
  }%
  \providecommand\rotatebox[2]{#2}%
  \ifx\svgwidth\undefined%
    \setlength{\unitlength}{300bp}%
    \ifx\svgscale\undefined%
      \relax%
    \else%
      \setlength{\unitlength}{\unitlength * \real{\svgscale}}%
    \fi%
  \else%
    \setlength{\unitlength}{\svgwidth}%
  \fi%
  \global\let\svgwidth\undefined%
  \global\let\svgscale\undefined%
  \makeatother%
  \begin{picture}(1,0.96110613)%
    \put(0,0){\includegraphics[width=\unitlength]{risk_ind_ROPT_IIDM.pdf}}%
    \put(0.42767976,0.10291233){\makebox(0,0)[lb]{\smash{0.5}}}%
    \put(0.39441725,0.10291233){\makebox(0,0)[lb]{\smash{1}}}%
    \put(0.30696099,0.10291233){\makebox(0,0)[lb]{\smash{2}}}%
    \put(0.21850472,0.10291233){\makebox(0,0)[lb]{\smash{3}}}%
    \put(0.13004845,0.10291233){\makebox(0,0)[lb]{\smash{4}}}%
    \put(0.28238663,0.08225906){\makebox(0,0)[lb]{\smash{$p$}}}%
    \put(0.07397842,0.14829367){\makebox(0,0)[lb]{\smash{0}}}%
    \put(0.07397842,0.22476295){\makebox(0,0)[lb]{\smash{2}}}%
    \put(0.07397842,0.30123223){\makebox(0,0)[lb]{\smash{4}}}%
    \put(0.07397842,0.37770152){\makebox(0,0)[lb]{\smash{6}}}%
    \put(0.04875878,0.22223369){\rotatebox{90}{\makebox(0,0)[lb]{\smash{$\bar{d}_\text{front,min}$ [m]}}}}%
    \put(0.4259432,0.5137951){\makebox(0,0)[lb]{\smash{0.5}}}%
    \put(0.3926807,0.5137951){\makebox(0,0)[lb]{\smash{1}}}%
    \put(0.30522443,0.5137951){\makebox(0,0)[lb]{\smash{2}}}%
    \put(0.21676816,0.5137951){\makebox(0,0)[lb]{\smash{3}}}%
    \put(0.130531188,0.5137951){\makebox(0,0)[lb]{\smash{4}}}%
    \put(0.28265007,0.49091872){\makebox(0,0)[lb]{\smash{$p$}}}%
    \put(0.07365433,0.54850386){\makebox(0,0)[lb]{\smash{0}}}%
    \put(0.05665433,0.63183578){\makebox(0,0)[lb]{\smash{10}}}%
    \put(0.05665433,0.71516769){\makebox(0,0)[lb]{\smash{20}}}%
    \put(0.05665433,0.79249959){\makebox(0,0)[lb]{\smash{30}}}%
    \put(0.0378952,0.63022801){\rotatebox{90}{\makebox(0,0)[lb]{\smash{$\bar{d}_\text{back,min}$ [m]}}}}%
    \put(0.14772334,0.89969559){\makebox(0,0)[lb]{\smash{IIDM: crash rate $\unit[37]{\%}$}}}%
    \put(0.60198282,0.10291233){\makebox(0,0)[lb]{\smash{0.1}}}%
    \put(0.64625603,0.10291233){\makebox(0,0)[lb]{\smash{1}}}%
    \put(0.68048666,0.10291233){\makebox(0,0)[lb]{\smash{2}}}%
    \put(0.77817857,0.10291233){\makebox(0,0)[lb]{\smash{5}}}%
    \put(0.9363688,0.10291233){\makebox(0,0)[lb]{\smash{10}}}%
    \put(0.7335003,0.06989007){\makebox(0,0)[lb]{\smash{$b^t$ $\unit[]{[\text{\euro}/km]}$}}}%
    \put(0.57262256,0.14829367){\makebox(0,0)[lb]{\smash{0}}}%
    \put(0.57262256,0.22476295){\makebox(0,0)[lb]{\smash{2}}}%
    \put(0.57262256,0.30123223){\makebox(0,0)[lb]{\smash{4}}}%
    \put(0.57262256,0.37770152){\makebox(0,0)[lb]{\smash{6}}}%
    \put(0.54440292,0.22223369){\rotatebox{90}{\makebox(0,0)[lb]{\smash{$\bar{d}_\text{front,min}$ [m]}}}}%
    \put(0.60298282,0.5137951){\makebox(0,0)[lb]{\smash{0.1}}}%
    \put(0.64525603,0.5137951){\makebox(0,0)[lb]{\smash{1}}}%
    \put(0.67788666,0.5137951){\makebox(0,0)[lb]{\smash{2}}}%
    \put(0.77817857,0.5137951){\makebox(0,0)[lb]{\smash{5}}}%
    \put(0.9363688,0.5137951){\makebox(0,0)[lb]{\smash{10}}}%
    \put(0.7335003,0.4841075){\makebox(0,0)[lb]{\smash{$b^t$ $\unit[]{[\text{\euro}/km]}$}}}%
    \put(0.13799947,0.01306426){\makebox(0,0)[lb]{\smash{$\lambda \hspace{-0.06cm} = \hspace{-0.06cm} \unit[5]{s}$}}}%
    \put(0.32062765,0.01306426){\makebox(0,0)[lb]{\smash{$\lambda\hspace{-0.06cm}= \hspace{-0.06cm}\unit[4]{s}$}}}%
    \put(0.49945535,0.01306426){\makebox(0,0)[lb]{\smash{$\lambda\hspace{-0.06cm}= \hspace{-0.06cm}\unit[3]{s}$}}}%
    \put(0.67776862,0.01306426){\makebox(0,0)[lb]{\smash{$\lambda\hspace{-0.06cm}=\hspace{-0.06cm} \unit[2]{s}$}}}%
    \put(0.8449138,0.01306426){\makebox(0,0)[lb]{\smash{bound}}}%
    \put(0.57280984,0.54650386){\makebox(0,0)[lb]{\smash{0}}}%
    \put(0.55552232,0.62383578){\makebox(0,0)[lb]{\smash{10}}}%
    \put(0.55552232,0.70716769){\makebox(0,0)[lb]{\smash{20}}}%
    \put(0.55552232,0.79049959){\makebox(0,0)[lb]{\smash{30}}}%
    \put(0.53846319,0.63022801){\rotatebox{90}{\makebox(0,0)[lb]{\smash{$\bar{d}_\text{back,min}$ [m]}}}}%
    \put(0.64977149,0.89969559){\makebox(0,0)[lb]{\smash{ROPT: crash rate $\unit[0]{\%}$}}}%
  \end{picture}%
\endgroup%

%% file: img/util_ind_ROPT_IIDM4.pdf_tex
\begingroup%
  \makeatletter%
  \providecommand\color[2][]{%
    \errmessage{(Inkscape) Color is used for the text in Inkscape, but the package 'color.sty' is not loaded}%
    \renewcommand\color[2][]{}%
  }%
  \providecommand\transparent[1]{%
    \errmessage{(Inkscape) Transparency is used (non-zero) for the text in Inkscape, but the package 'transparent.sty' is not loaded}%
    \renewcommand\transparent[1]{}%
  }%
  \providecommand\rotatebox[2]{#2}%
  \ifx\svgwidth\undefined%
    \setlength{\unitlength}{300.0bp}%
    \ifx\svgscale\undefined%
      \relax%
    \else%
      \setlength{\unitlength}{\unitlength * \real{\svgscale}}%
    \fi%
  \else%
    \setlength{\unitlength}{\svgwidth}%
  \fi%
  \global\let\svgwidth\undefined%
  \global\let\svgscale\undefined%
  \makeatother%
  \begin{picture}(1,0.96110613)%
    \put(0,0){\includegraphics[width=\unitlength]{util_ind_ROPT_IIDM4.pdf}}%
    \put(0.42867976,0.10401233){\makebox(0,0)[lb]{\smash{0.5}}}%
    \put(0.39441725,0.10401233){\makebox(0,0)[lb]{\smash{1}}}%
    \put(0.30696099,0.10401233){\makebox(0,0)[lb]{\smash{2}}}%
    \put(0.21850472,0.10401233){\makebox(0,0)[lb]{\smash{3}}}%
    \put(0.13304845,0.10401233){\makebox(0,0)[lb]{\smash{4}}}%
    \put(0.28738663,0.08075906){\makebox(0,0)[lb]{\smash{$p$}}}%
    \put(0.07337842,0.14629367){\makebox(0,0)[lb]{\smash{4}}}%
    \put(0.07337842,0.24376295){\makebox(0,0)[lb]{\smash{5}}}%
    \put(0.07337842,0.33963223){\makebox(0,0)[lb]{\smash{6}}}%
    \put(0.07337842,0.43470152){\makebox(0,0)[lb]{\smash{7}}}%
    \put(0.04575878,0.27223369){\rotatebox{90}{\makebox(0,0)[lb]{\smash{$\bar{t}_\text{gap}$ [s]}}}}%
    \put(0.42867976,0.5127951){\makebox(0,0)[lb]{\smash{0.5}}}%
    \put(0.39441725,0.5127951){\makebox(0,0)[lb]{\smash{1}}}%
    \put(0.30696099,0.5127951){\makebox(0,0)[lb]{\smash{2}}}%
    \put(0.21850472,0.5127951){\makebox(0,0)[lb]{\smash{3}}}%
    \put(0.13304845,0.5127951){\makebox(0,0)[lb]{\smash{4}}}%
    \put(0.28565007,0.49291872){\makebox(0,0)[lb]{\smash{$p$}}}%
    \put(0.07437842,0.54250386){\makebox(0,0)[lb]{\smash{1}}}%
    \put(0.07437842,0.62883578){\makebox(0,0)[lb]{\smash{3}}}%
    \put(0.07437842,0.72616769){\makebox(0,0)[lb]{\smash{5}}}%
    \put(0.07437842,0.81949959){\makebox(0,0)[lb]{\smash{7}}}%
    \put(0.04575878,0.68022801){\rotatebox{90}{\makebox(0,0)[lb]{\smash{$\bar{n}_\text{gap}$}}}}%
    \put(0.14772334,0.90619559){\makebox(0,0)[lb]{\smash{IIDM: crash rate $\unit[37]{\%}$}}}%
    \put(0.60098282,0.10401233){\makebox(0,0)[lb]{\smash{0.1}}}%
    \put(0.64625603,0.10401233){\makebox(0,0)[lb]{\smash{1}}}%
    \put(0.67848666,0.10401233){\makebox(0,0)[lb]{\smash{2}}}%
    \put(0.78017857,0.10401233){\makebox(0,0)[lb]{\smash{5}}}%
    \put(0.9383688,0.10401233){\makebox(0,0)[lb]{\smash{10}}}%
    \put(0.7335003,0.07049007){\makebox(0,0)[lb]{\smash{$b^t$ $\unit[]{[\text{\euro}/km]}$}}}%
    \put(0.57162256,0.14729367){\makebox(0,0)[lb]{\smash{4}}}%
    \put(0.57162256,0.24376295){\makebox(0,0)[lb]{\smash{5}}}%
    \put(0.57162256,0.33963223){\makebox(0,0)[lb]{\smash{6}}}%
    \put(0.57162256,0.43570152){\makebox(0,0)[lb]{\smash{7}}}%
    \put(0.54640292,0.27223369){\rotatebox{90}{\makebox(0,0)[lb]{\smash{$\bar{t}_\text{gap}$ [s]}}}}%
    \put(0.60298282,0.5127951){\makebox(0,0)[lb]{\smash{0.1}}}%
    \put(0.64825603,0.5127951){\makebox(0,0)[lb]{\smash{1}}}%
    \put(0.68048666,0.5127951){\makebox(0,0)[lb]{\smash{2}}}%
    \put(0.78217857,0.5167951){\makebox(0,0)[lb]{\smash{5}}}%
    \put(0.9423688,0.5127951){\makebox(0,0)[lb]{\smash{10}}}%
    \put(0.7335003,0.4871075){\makebox(0,0)[lb]{\smash{$b^t$ $\unit[]{[\text{\euro}/km]}$}}}%
    \put(0.22099947,0.01366426){\makebox(0,0)[lb]{\smash{$\lambda \hspace{-0.06cm} = \hspace{-0.06cm} \unit[5]{s}$}}}%
    \put(0.40262765,0.01366426){\makebox(0,0)[lb]{\smash{$\lambda\hspace{-0.06cm}= \hspace{-0.06cm}\unit[4]{s}$}}}%
    \put(0.58345535,0.01366426){\makebox(0,0)[lb]{\smash{$\lambda\hspace{-0.06cm}= \hspace{-0.06cm}\unit[3]{s}$}}}%
    \put(0.76076862,0.01366426){\makebox(0,0)[lb]{\smash{$\lambda\hspace{-0.06cm}=\hspace{-0.06cm} \unit[2]{s}$}}}%
    \put(0.57162256,0.54250386){\makebox(0,0)[lb]{\smash{1}}}%
    \put(0.57162256,0.62883578){\makebox(0,0)[lb]{\smash{3}}}%
    \put(0.57162256,0.72416769){\makebox(0,0)[lb]{\smash{5}}}%
    \put(0.57162256,0.81949959){\makebox(0,0)[lb]{\smash{7}}}%
    \put(0.54640292,0.68022801){\rotatebox{90}{\makebox(0,0)[lb]{\smash{$\bar{n}_\text{gap}$}}}}%
    \put(0.64977149,0.90619559){\makebox(0,0)[lb]{\smash{ROPT: crash rate $\unit[0]{\%}$}}}%
  \end{picture}%
\endgroup%

%% file: chapters/outlook.tex
\vspace{0.3cm}
\section{Conclusion and Outlook}
\label{sec:outlook}

In this work, we outlined a longitudinal velocity planner along map geometries. ROPT alternates between trajectory generation and evaluation in an optimization cycle. Multiple double-ramp velocity profiles as well as fixed constant velocity, acceleration and deceleration trajectories are constructed. For the arising dynamic traffic scenario predictions, the integral risk, utility and comfort are calculated. ROPT infers curve and collision risks based on Gaussian probability estimation and a future risk accumulation with the survival analysis. 
The cycle is repeated until the cost threshold is satisfied for each sample in the set and the velocity course with the lowest cost is chosen for execution.  

As a comparison, we extended the IDM for curve driving and intersection scenarios to obtain the IIDM. It uses a modified free term for maximal curvature adherence and lane change criterions altering the interaction term. 
In simulations of right turning at T-intersections with several cross traffic densities, ROPT considers the risks properly in all parameter settings and distributes it to the two risk sources evenly. IIDM has qualitatively similar results but can yield crash cases in suboptimal configurations, because of if-then type behavior and improper prediction. We thus created the predictive IIDM, which applies the heuristics for all time steps and has no accidents but weak risk-utility tradeoff.

\begin{figure}[t!]
  \centering
  \vspace{-0.46cm}
  \resizebox{1.0\linewidth}{!}{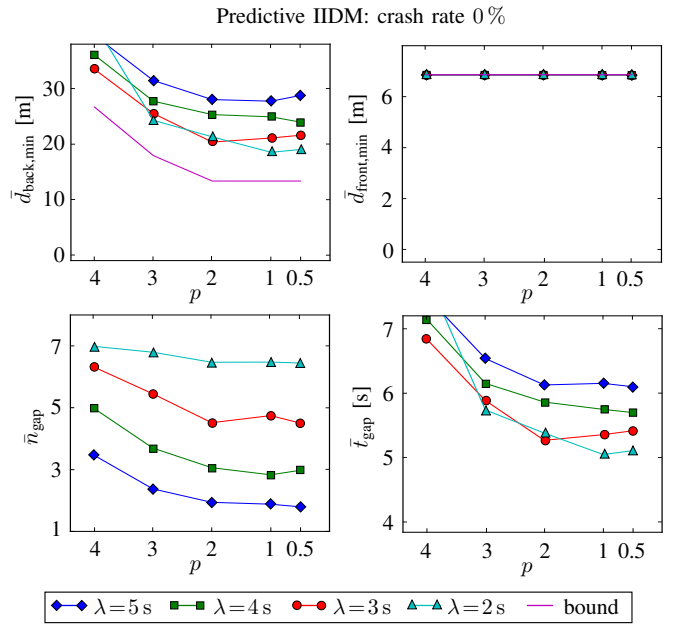}
  \caption{Predictive IIDM with risk and utility indicators.} 
  \label{fig:pred-idm}
\end{figure} 

Overall, ROPT is generalizable to more complex situations due to its holistic risk prediction. 
In \cite{intwarn2017}, we modeled occlusion risks from buildings around intersections with virtual cars.
In future work, traffic elements at intersections (e.g. traffic lights or right-before-left) could be incorporated in a similar way. Otherwise, the ego car will not comply with special rule-based behaviors. 

The survival method has been shown to detect collision risks early with few false positives \cite{eggert2017}.
Nevertheless for high prediction horizons, the 2D Gaussians might cover opposite lanes when turning and project errors. It remains to be investigated, if the road curvatures are usable to create Gaussian Mixture Models (GMM) with bent shapes. 

Vehicle dynamics are currently only modeled with a double integrator system in ROPT. For online application, physical steering models with delay of gas pedal to acceleration for the ego car and sensor models with noisy and late detected positions of other cars may be included as a pre-processing step. 
At last, the weighting of risk against benefit depends on the driver type. We expect that self-adjustment of the corresponding parameters would yield a personalization of planned velocites.

%% file: img/ris_and_util_pred_IDM.pdf_tex
\begingroup%
  \makeatletter%
  \providecommand\color[2][]{%
    \errmessage{(Inkscape) Color is used for the text in Inkscape, but the package 'color.sty' is not loaded}%
    \renewcommand\color[2][]{}%
  }%
  \providecommand\transparent[1]{%
    \errmessage{(Inkscape) Transparency is used (non-zero) for the text in Inkscape, but the package 'transparent.sty' is not loaded}%
    \renewcommand\transparent[1]{}%
  }%
  \providecommand\rotatebox[2]{#2}%
  \ifx\svgwidth\undefined%
    \setlength{\unitlength}{300.0bp}%
    \ifx\svgscale\undefined%
      \relax%
    \else%
      \setlength{\unitlength}{\unitlength * \real{\svgscale}}%
    \fi%
  \else%
    \setlength{\unitlength}{\svgwidth}%
  \fi%
  \global\let\svgwidth\undefined%
  \global\let\svgscale\undefined%
  \makeatother%
  \begin{picture}(1,0.99110613)%
    \put(0,0){\includegraphics[width=\unitlength]{ris_and_util_pred_IDM.pdf}}%
    \put(0.4229432,0.10091233){\makebox(0,0)[lb]{\smash{0.5}}}%
    \put(0.3896807,0.10091233){\makebox(0,0)[lb]{\smash{1}}}%
    \put(0.30222443,0.10091233){\makebox(0,0)[lb]{\smash{2}}}%
    \put(0.21376816,0.10091233){\makebox(0,0)[lb]{\smash{3}}}%
    \put(0.127531188,0.10091233){\makebox(0,0)[lb]{\smash{4}}}%
    \put(0.27765007,0.07925906){\makebox(0,0)[lb]{\smash{$p$}}}%
    \put(0.06897842,0.13029367){\makebox(0,0)[lb]{\smash{1}}}%
    \put(0.06897842,0.21976295){\makebox(0,0)[lb]{\smash{3}}}%
    \put(0.06897842,0.31423223){\makebox(0,0)[lb]{\smash{5}}}%
    \put(0.06897842,0.40470152){\makebox(0,0)[lb]{\smash{7}}}%
    \put(0.04875878,0.26823369){\rotatebox{90}{\makebox(0,0)[lb]{\smash{$\bar{n}_\text{gap}$}}}}%
    \put(0.4229432,0.5077951){\makebox(0,0)[lb]{\smash{0.5}}}%
    \put(0.3896807,0.5077951){\makebox(0,0)[lb]{\smash{1}}}%
    \put(0.30222443,0.5077951){\makebox(0,0)[lb]{\smash{2}}}%
    \put(0.21376816,0.5077951){\makebox(0,0)[lb]{\smash{3}}}%
    \put(0.127531188,0.5077951){\makebox(0,0)[lb]{\smash{4}}}%
    \put(0.27765007,0.487091872){\makebox(0,0)[lb]{\smash{$p$}}}%
    \put(0.07365433,0.54250386){\makebox(0,0)[lb]{\smash{0}}}%
    \put(0.05665433,0.62683578){\makebox(0,0)[lb]{\smash{10}}}%
    \put(0.05665433,0.71016769){\makebox(0,0)[lb]{\smash{20}}}%
    \put(0.05665433,0.79249959){\makebox(0,0)[lb]{\smash{30}}}%
    \put(0.0378952,0.63022801){\rotatebox{90}{\makebox(0,0)[lb]{\smash{$\bar{d}_\text{back,min}$ [m]}}}}%
    \put(0.31772334,0.90169559){\makebox(0,0)[lb]{\smash{Predictive IIDM: crash rate $\unit[0]{\%}$}}}%
    \put(0.62398282,0.09791233){\makebox(0,0)[lb]{\smash{4}}}%
    \put(0.71588666,0.09791233){\makebox(0,0)[lb]{\smash{3}}}%
    \put(0.80225603,0.09791233){\makebox(0,0)[lb]{\smash{2}}}%
    \put(0.89017857,0.09791233){\makebox(0,0)[lb]{\smash{1}}}%
    \put(0.9203688,0.09791233){\makebox(0,0)[lb]{\smash{0.5}}}%
    \put(0.7755003,0.07725906){\makebox(0,0)[lb]{\smash{$p$}}}%
    \put(0.56752232,0.14029367){\makebox(0,0)[lb]{\smash{4}}}%
    \put(0.56752232,0.23776295){\makebox(0,0)[lb]{\smash{5}}}%
    \put(0.56752232,0.33323223){\makebox(0,0)[lb]{\smash{6}}}%
    \put(0.56752232,0.42970152){\makebox(0,0)[lb]{\smash{7}}}%
    \put(0.54440292,0.26323369){\rotatebox{90}{\makebox(0,0)[lb]{\smash{$\bar{t}_\text{gap}$ [s]}}}}%
    \put(0.62298282,0.5057951){\makebox(0,0)[lb]{\smash{4}}}%
    \put(0.70988666,0.5057951){\makebox(0,0)[lb]{\smash{3}}}%
    \put(0.80025603,0.5057951){\makebox(0,0)[lb]{\smash{2}}}%
    \put(0.88817857,0.5057951){\makebox(0,0)[lb]{\smash{1}}}%
    \put(0.9183688,0.5057951){\makebox(0,0)[lb]{\smash{0.5}}}%
    \put(0.7755003,0.487091872){\makebox(0,0)[lb]{\smash{$p$}}}%
    \put(0.13299947,0.01376426){\makebox(0,0)[lb]{\smash{$\lambda \hspace{-0.06cm} = \hspace{-0.06cm} \unit[5]{s}$}}}%
    \put(0.31262765,0.01376426){\makebox(0,0)[lb]{\smash{$\lambda\hspace{-0.06cm}= \hspace{-0.06cm}\unit[4]{s}$}}}%
    \put(0.49245535,0.01376426){\makebox(0,0)[lb]{\smash{$\lambda\hspace{-0.06cm}= \hspace{-0.06cm}\unit[3]{s}$}}}%
    \put(0.66376862,0.01376426){\makebox(0,0)[lb]{\smash{$\lambda\hspace{-0.06cm}=\hspace{-0.06cm} \unit[2]{s}$}}}%
    \put(0.8389138,0.01376426){\makebox(0,0)[lb]{\smash{bound}}}%
    \put(0.56752232,0.54850386){\makebox(0,0)[lb]{\smash{0}}}%
    \put(0.56552232,0.62483578){\makebox(0,0)[lb]{\smash{2}}}%
    \put(0.56552232,0.70416769){\makebox(0,0)[lb]{\smash{4}}}%
    \put(0.56552232,0.78049959){\makebox(0,0)[lb]{\smash{6}}}%
    \put(0.53846319,0.63022801){\rotatebox{90}{\makebox(0,0)[lb]{\smash{$\bar{d}_\text{front,min}$ [m]}}}}%
  \end{picture}%
\endgroup%